\documentclass{vgtc}                          % final (conference style)
\ifpdf%                                % if we use pdflatex
  \pdfoutput=1\relax                   % create PDFs from pdfLaTeX
  \usepackage{graphicx}                % allow us to embed graphics files
  \DeclareGraphicsExtensions{.pdf,.png,.jpg,.jpeg} % for pdflatex we expect .pdf, .png, or .jpg files
\else%                                 % else we use pure latex
  \usepackage{graphicx}                % allow us to embed graphics files
  \DeclareGraphicsExtensions{.eps}     % for pure latex we expect eps files
\fi%

\graphicspath{{figures/}{pictures/}{images/}{./}} % where to search for the images

\usepackage{microtype}                 % use micro-typography (slightly more compact, better to read)
\PassOptionsToPackage{warn}{textcomp}  % to address font issues with \textrightarrow
\usepackage{textcomp}                  % use better special symbols
\usepackage{mathptmx}                  % use matching math font
\usepackage{times}                     % we use Times as the main font
\usepackage{cite}                      % needed to automatically sort the references
\usepackage{tabu}                      % only used for the table example
\usepackage{booktabs}                  % only used for the table example
\usepackage[usenames, dvipsnames]{color}

\newcommand{\clrr}{\textcolor{black}}
\newcommand{\clrb}{\textcolor{black}}

\onlineid{1176}

\vgtccategory{algorithm/technique}

\vgtcinsertpkg

\title{Relationship-aware Multivariate Sampling Strategy for \clrb{Scientific Simulation Data}}

\author{Subhashis Hazarika\thanks{e-mail list: [shazarika, ayan, pwolfram, earl, nurban]@lanl.gov} %
\and Ayan Biswas %
\and Phillip J. Wolfram %
\and Earl Lawrence %
\and Nathan Urban
}
\affiliation{\scriptsize Los Alamos National Laboratory, New Mexico, USA}

\abstract{With the increasing computational power of current supercomputers, the size of data produced by scientific simulations is rapidly growing. To reduce the storage footprint and facilitate scalable post-hoc analyses of such scientific data sets, various data reduction/summarization methods have been proposed over the years. Different flavors of sampling algorithms exist to sample the high-resolution scientific data, while preserving important data properties required for subsequent analyses. However, most of these sampling algorithms are designed for univariate data and cater to post-hoc analyses of single variables. In this work, we propose a multivariate sampling strategy which preserves the original variable relationships and enables different multivariate analyses directly on the sampled data. Our proposed strategy utilizes \textit{principal component analysis} to capture the variance of multivariate data and can be built on top of any existing state-of-the-art sampling algorithms for single variables. In addition, we also propose variants of different data partitioning schemes (regular and irregular) to efficiently model the local multivariate relationships. Using two real-world multivariate data sets, we demonstrate the efficacy of our proposed multivariate sampling strategy with respect to its data reduction capabilities as well as the ease of performing efficient post-hoc multivariate analyses. 
} % end of abstract

\begin{document}


\firstsection{Introduction}

\maketitle

Scientists frequently simulate multiple physical variables/attributes at the same time in their computational models.
The resulting multivariate simulation data is analyzed to understand the variable relationships and how they interact to influence the simulated physical phenomenon. 
%Recent advancements in the field of computational sciences have enabled scientists to design high-resolution simulation models producing large-scale scientific data sets. 
\clrb{However, with increasing data size, it is becoming computationally prohibitive and challenging to analyze and visualize such high-dimensional simulation data.}
%Given the high-dimensional nature of multivariate data, it is even more challenging to perform essential multivariate analyses for large-scale multivariate simulation data.

%the problem is even more challenging for large-scale multivariate simulation data.

A popular and effective strategy to address these challenges is to reduce the data size by sampling important features of the data while data still resides in the memory~\cite{insitu_STAR,Hank_SC_2015}. Instead of storing the high-resolution data sets, the corresponding sampled data is stored and subsequently used for various post-hoc analyses. Different flavors of sampling algorithms exist in literature~\cite{ayan_1dsampling, ayan_2dsampling, void_cluster_sampling,Dutta_mvsample_2019} which can selectively sample different data properties. However, most of these algorithms primarily target univariate data. 
%or a specific derived property of multivariate data. 
To perform traditional  multivariate analyses such as correlation studies between variables and joint multivariate queries across different variables directly on the sampled data, it is important to preserve the variable relationships while sampling the data. Using the univariate sampling methods to sample the multivariate datasets can potentially fail to preserve the important inter-variable relationships, and the subsequent post-hoc multivariate analyses can become unreliable. \clrb{Further, given the correlation existing across the variables, it is not necessary to explicitly store all the variables}. Therefore, it is possible to achieve much larger data reduction for multivariate datasets if we consider the variable relationships while sampling the data.   

%Further, correlations existing across the variables can help achieve much larger data reduction since not all of the correlated variables need to be stored during in-situ processing. [one more sentence to finish the thought?] 

In this paper, we propose a multivariate sampling strategy to preserve and utilize the original multivariate relationships to facilitate higher data reduction as well as enable an efficient post-hoc exploration workflow. To efficiently model the complex global non-linear multivariate relationships, we use a locally piece-wise linear model~\cite{localpca,locallinearembedding} by first partitioning the spatial domain. \clrb{In this regard, we propose variants of different partitioning schemes (regular and irregular), especially adapted for multivariate data.}
The local linear relationship for each partition is modeled using Principal Component Analysis (PCA). Using PCA, we extract the correlations among the variables and proceed to achieve data reduction by selecting an optimal number of uncorrelated variables. We further reduce the data footprint by sampling the spatial domain in that uncorrelated variable space. PCA also provides the means to perform uncertainty quantification that can be controlled by the user as an error-tolerance. 

%\textbf{[This paragraph should be smoothed out to blend in with the previous one]}
%While modeling the global multivariate relationship using locally linear models it is important to partition the data efficiently. In this regard, we propose variants of different partitioning schemes (regular and irregular), adapted for multivariate data. 
%We also adapt different partitioning schemes (regular and irregular) to model the local multivariate relationships in a better way using local PCA models across the spatial domain. 
%\clrb{This also helps us apply the proposed sampling strategy independently on individual partitions, which is often the setup for large-scale simulations in distributed computing environments.} 
\clrb{To demonstrate the efficacy of our proposed strategy, we apply it on a two-dimensional ocean simulation data set with 75 variables and a three-dimensional hurricane data set with 13 variables.} We used three different partitioning schemes and two popular sampling algorithms to highlight its compatibility with existing sampling and partitioning methods. To summarize, the main contribution of our work is threefold:

\begin{enumerate}
    \setlength\itemsep{0.001em}
	\item We formulate a sampling strategy for large-scale multivariate data which utilizes the intrinsic redundancy of the variables and the spatial dimensions to achieve larger data reduction. 
	
	%utilizes variable relationships to reduce variable dimensions and various sampling algorithms to reduce the spatial dimensions for large-scale multivariate data.
	\item We facilitate various post-hoc multivariate analyses directly on the reduced/sampled data, without the need to reconstruct the high-resolution scalar fields for all the variables.
	\item We propose multivariate relationship-aware spatial domain decomposition schemes to extract the locally linear models. 
\end{enumerate}   

%For each local partition of the spatial domain, we first perform a PCA and identify the minimum number of principal components (PCs) required to capture a user desired variance of the data. For example, in a multivariate system with $D$ variables, we preserve $d$ ($d < D$) dimensions (PCs) required to capture $99\%$ variance of the data. This significantly reduces the variables dimensions that we need to deal with without significantly compromising on the variable relationship. Next different sampling algorithms can be applied on the first PC field, which essentially captures the maximum variance of the multivariate data. The final samples can be utilized to perform various post-hoc multivariate analysis and visualization tasks in a scalable manner.    

\section{Related Work}
\textbf{Sampling-based Data Analysis:} Data sampling methods have been widely used in the visualization community to reduce the size of large-scale data sets in order to facilitate timely execution of various visualization and analysis activities. To enable interactive visualization of large-scale cosmology simulation data, Woodring et al.~\cite{Woodring_sampling} proposed a stratified random sampling approach. 
%Park et al.~\cite{park16} proposed visualization-aware sampling methods, particularly targeting scatter-plot and map-plot visualizations. 
Nguyen and Song~\cite{nguyen16} incorporated centrality-driven clustering information during random sampling. Using the ideas of entropy maximization, Biswas et al.~\cite{ayan_1dsampling, ayan_2dsampling} recently proposed \textit{in situ} data-driven sampling schemes that preserve important data features along with their gradient properties. For scattered datasets, Rapp et al.~\cite{void_cluster_sampling} proposed a blue noise preserving sampling method to identify representative subset of points. However, these sampling methods are primarily targeted for univariate data fields or a very specific derived property of the multivariate data. For instance, Dutta et al.~\cite{Dutta_mvsample_2019} recently proposed a pointwise mutual information based approach for multivariate sampling to identify regions with high mutual information among the variables. \clrb{In this paper, we preserve the overall variable relationships to enable more generic multivariate analyses directly on the sampled data.} 

%For the purpose of more generic multivariate analyses, though, the overall variable relationship has to be preserved across the samples. To the best of our knowledge, no current sampling-base visualization method preserves the variable relationships and therefore cannot facilitate generic multivariate analyses on the sampled data. 

\textbf{Multivariate Data Analysis:} This is an important area of research in the scientific visualization community. To visualize multivariate relationship across the spatial domain Sauber et al.~\cite{Sauber2006} analyzed the correlation coefficients among the variables in local neighborhoods. Correlation analysis was also extended to enable different query-driven methods for multivariate data ~\cite{Bethel2007}. Gosnik et al~\cite{gosink_dist_qdv} improved multivariate query-driven analysis by using various statistical models. To identify interesting regions in multivariate data, J\"{a}nicke et al.~\cite{Janicke2007} adapted different local statistical measures quantifying variable information. 
%Biswas et al.~\cite{BiswasDSW13} proposed an information-theoretic framework to quantify variable importance and understand their relationships.
For large-scale multivariate simulation data, Hazarika et al.~\cite{codda18} proposed copula functions~\cite{coupla17} to model variable relationship \textit{in situ} along with different statistical distribution models to reduce storage footprint. For a more extensive review of multivariate data analysis and visualization research readers can refer to Wong et al.~\cite{Wong1994} and Fuchs et al.~\cite{Fuchs}. 
%For a more extensive review of multivariate data analysis and visualization research readers can refer to the detail studies performed by Wong et al.~\cite{Wong1994} and Fuchs et al.~\cite{Fuchs}. 
%In this work, we enable such multivariate analyses directly on the sampled data obtained via sampling algorithms.

\section{Method}
\begin{figure}[t!]
\centering
        \includegraphics[width = 0.99\columnwidth]{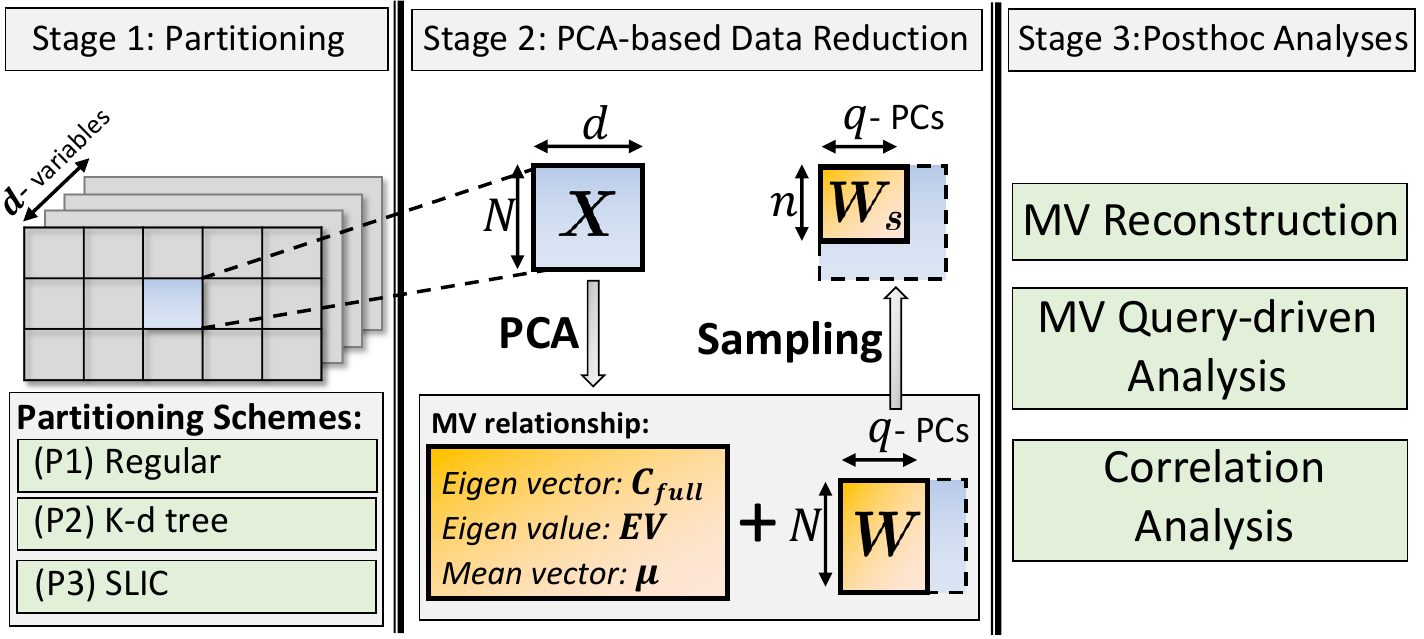}
\clrb{\caption{A high-level illustration of the different stages of our proposed multivariate sampling strategy.}}
\label{overview}
\end{figure}

\clrb{\textbf{Overview:} Fig.~\ref{overview} provides a high-level-illustration of our proposed strategy. To efficiently model the multivariate relationships using locally linear models, the spatial domain is first decomposed into smaller partitions using three different partitioning schemes. Next, for each partition, we apply PCA and sampling algorithms to reduce the overall storage footprint of multivariate data while preserving the variable relationships. Finally, we perform different post-hoc multivariate analyses directly on the reduced/sampled data.} 

\clrb{In this paper, we first introduce the concept of PCA and its properties for multivariate relationship modeling in subsection 3.1. In subsection 3.2, we discuss the different sampling algorithms for spatial data reduction. Despite being the first step in our workflow, we leave the detail discussion of the multivariate partitioning schemes to subsection 3.3, after certain notations and concepts related to PCA have been explained in subsection 3.1. The various post-hoc analyses are covered in Section 4.}     

% In our proposed approach, to efficiently model the multivariate relationships using locally linear models, the spatial domain is first decomposed into smaller partitions. For each partition, we apply PCA and sampling algorithms to reduce the overall storage footprint of multivariate data while preserving the variable relationship. Fig.~\ref{overview} provides a high-level-illustration of our proposed strategy. 

%For each partition, using PCA, the correlated variables are transformed into uncorrelated variables and an optimal number of variables are stored. Sampling algorithms are applied on the transformed variables to further reduce the data size.  

%The primary goal of the proposed multivariate sampling strategy is to preserve the original variable relationships while reducing the overall storage footprint, via sampling. To achieve this, we first apply PCA on individual spatial partitions to reduce the number of variables (dimensions), capturing a user-specified maximum variance of the original variables. Next, we apply different sampling algorithms to reduce the number of datapoints to store in each partition. Fig.~\ref{overview} provides a high-level illustration of our approach. 

%In this section, we first discuss in detail the idea behind the proposed multivariate sampling strategy and then elaborate on the different partitioning schemes used in our work.

\subsection{Variable Dimension Reduction using PCA}
Multivariate data comprises of multiple interrelated variables with different degrees of association among them. The central idea of PCA is to project these variables to a new set of uncorrelated variables/dimensions, called \textit{principal components} (PC's), which are ordered in such a way that the first few retain most of the variation in all of the original variables. This property of PCA is useful to quantitatively decide how many dimensions (PCs) to store to capture a given fraction of the variation of the original data. We apply this dimensionality reduction property of PCA to identify the number of PC's to store for individual partitions of the data. 

As illustrated in Fig.~\ref{overview}, consider a partition with $N$ datapoints and $d$ variables. Let $\mathbf{X}$ ($N \times d$ matrix) denote the multivariate data in this partition. Then $\mathbf{X}$ can be expressed by the following linear combination,
\begin{equation}
    \mathbf{X} = \mathbf{WC_q} + \mathbf{\mu} + \epsilon
\label{pca_e1}
\end{equation}
where, the new basis $\mathbf{C_q}$ ($q \times d$ matrix) represents the top $q$ PC's that capture the maximum variation of $\mathbf{X}$. $\mathbf{W}$ ($N \times q$ matrix) represents the projections of the original $N$ datapoints onto these $q$ orthogonal directions and $\mathbf{\mu}$ represents  the $d$-dimensional mean vector of $\mathbf{X}$. $\epsilon$ is the residual error associated with the loss of dimensions (i.e, $q < d$). When number of PC's $q = d$, $\epsilon = 0$. Therefore, given $\mathbf{W}$, $\mathbf{C_q}$ and $\mathbf{\mu}$, we can approximate the original multivariate data $\mathbf{X}$ using Equation~\ref{pca_e1}. Eigen decomposition of the covariance matrix of $\mathbf{X}$ (i.e, $\frac{1}{N-1}\mathbf{X^T}\mathbf{X}$) gives the full $d \times d$ PC matrix $\mathbf{C_{full}}$ (aka. eigen vectors) and their corresponding explained variances $\mathbf{EV}$ (aka. eigen values). The first $q$ PC's in $\mathbf{C_{full}}$ makes up the matrix $\mathbf{C_q}$. The transformed data $\mathbf{W}$ is obtained as follows:
\begin{equation}
    \mathbf{W} = (\mathbf{X - \mu})\mathbf{C_q}^T
\label{pca_e2}
\end{equation}
where, $\mathbf{C_q}^T$ is the transpose of $\mathbf{C_q}$. 

\textbf{Error Quantification:} An advantage of using PCA to reduce variable dimensions is that we can estimate a threshold on the reconstruction error of the original variables in the post-hoc analysis phase. This is because in PCA maximizing variance is equivalent to minimizing the residual error $\epsilon$ of reconstructing the original data back. This relationship between explained variance of the PC's with the residual error (least-square error) is given as,
\begin{equation}
    Explained\_Variance\_Ratio = 1 - \frac{\epsilon^2}{\| \mathbf{X} \|^2}
\label{pca_e3}
\end{equation}
where, $\epsilon^2 = \| \mathbf{X} - (\mathbf{WC + \mu}) \|^2$ is the squared norm of the residual error. The term $\frac{\epsilon^2}{\| \mathbf{X} \|^2}$ corresponds to the normalized residual error corresponding to the original data and is inversely proportional to the percentage of variance captured by the PC's. \clrb{For each partition, given the desired maximum variance required to preserve, we can decide the optimal number of PC's $q$ ($<d$) to store.}
%For each partition, after we decide on the optimal number of PC's to store (i.e, $q$) we store the corresponding PCA meta-data $\mathbf{C}$, $\mu$ and the corresponding explained variances $EV$ (i.e, eigen values) required to successfully perform post-hoc multivariate analyses. 

%After reducing the variable dimension, sampling algorithms are applied on the transformed data $\mathbf{W}$ to reduce the number of datapoints to eventually store for each partition.

\begin{figure}[t!]
\centering

        \includegraphics[width = 0.95\columnwidth]{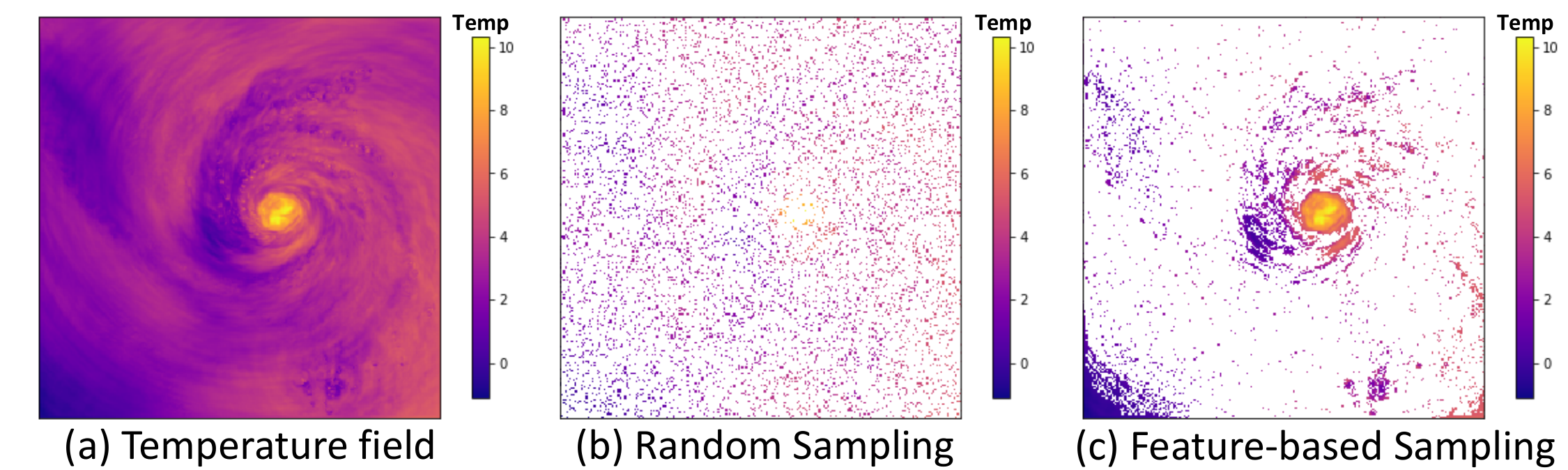} 
\caption{Univariate sampling algorithm results: (a) Temperature field of Isabel dataset. (b) 5$\%$ random sampling. (c) 5$\%$ feature-based sampling~\cite{ayan_1dsampling} which gives more importance to the data features. }
\label{sampling_fig}
\end{figure}

% $\mathbf{C}$ is a $q \times d$ matrix representing the top $q$ eigen vectors of the covariance matrix, $\mathbf{X}$. These $q$ eigenvectors represent the $q$ principal components (directions) along which the variance is maximized. The corresponding variance captured by these PC's is given by their eigenvalues. $W$ is a $N \times q$ matrix representing the projects of the original $N$ datapoints onto these $q$ orthogonal directions.  

\subsection{Spatial Data Reduction using Sampling Algorithms}
%The transformed data $\mathbf{W}$ is an $N \times q$ matrix with $N$ datapoints and $q$ uncorrelated variables along the $q$ orthogonal PC's. 
After reducing the variable dimensions from $d$ to $q$ using PCA, we next decide how many datapoints to retain out of $N$ using sampling algorithms on the uncorrelated variable. In our work, to obtain consistent samples for all variables, we apply the sampling algorithms only on the scalar field of the first PC, which captures the maximum variation of the original multivariate data. This helps us to efficiently perform post-hoc multivariate analyses on the sampled data. Depending on the analysis requirements, different state-of-the-art sampling algorithms can be utilized to perform spatial data reduction. In this work, we consider the following two distinct flavors of sampling algorithms commonly used for scientific data sets.

\textbf{(S1) Random Sampling:} Random sampling is a popular sampling technique for data summarization because the sampled data points preserve the original data distribution along with statistical properties like mean, standard deviation etc. It can also be readily combined with other importance-based sampling methods to induce randomness in the samples.

\textbf{(S2) Feature-based Sampling:} Scientific data sets often contain important low-frequency data features which can get eliminated with random sampling. Recently, Biswas et al.~\cite{ayan_1dsampling} proposed an importance sampling algorithm using the idea of entropy maximization to preserve important local data features in the sampled data. 

% \begin{itemize}
%     \item \textbf{(S1) Random Sampling:} Random sampling is a popular sampling technique for data summarization because the sampled data points preserve the original data distribution along with statistical properties like mean, standard deviation etc. It can also be readily combined with other importance-based sampling methods to induce randomness in the samples.
%     %During random sampling each data point has an equal probability of being picked and hence is independent of the variable values.
    
%     \item \textbf{(S2) Feature-based Sampling:} Scientific data sets often contain important low-frequency data features which can get eliminated with random sampling. Recently, Biswas et al.~\cite{ayan_1dsampling} proposed an importance sampling algorithm using the idea of entropy maximization to preserve important local data features in the sampled data. 
% \end{itemize}

In Figure~\ref{sampling_fig}, using a slice of the Temperature field of the Hurricane Isabel data set, we demonstrate the properties of the two sampling algorithms. As can be seen in Fig.~\ref{sampling_fig}(b), with random sampling, each data point has an equal probability of being picked, irrespective of the data features. Whereas, feature-based sampling algorithm (Fig.~\ref{sampling_fig}c) puts more emphasis on the intrinsic data features (i.e, hurricane eye). 
%It is also possible to combine both methods to have flavors of both the algorithms in the eventual samples.     

\begin{figure}[t!]
\centering
        \includegraphics[width = 0.98\columnwidth]{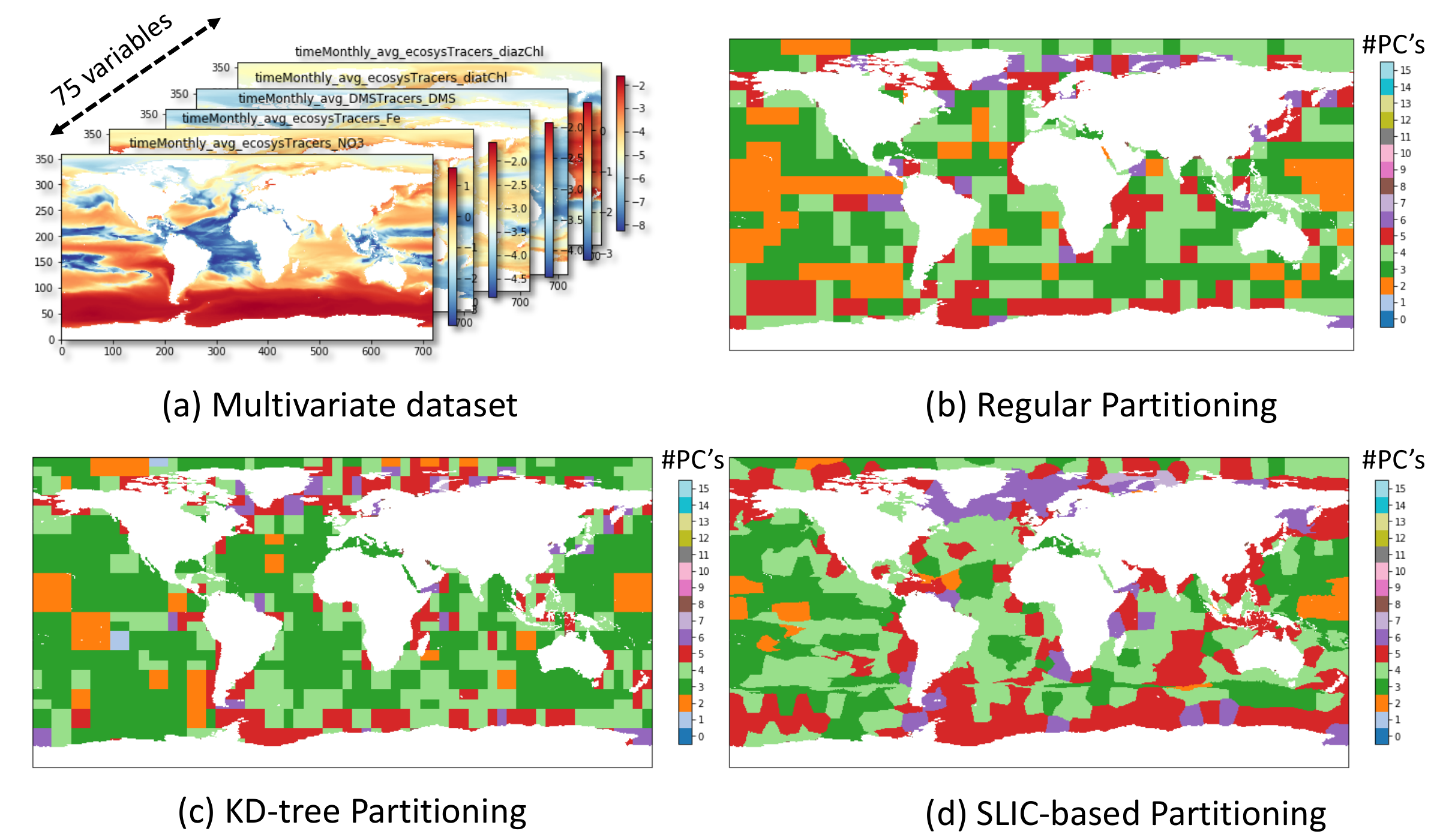} 
\caption{Optimal number of components (PC's) required for each partition to capture 99$\%$ variance of the multivariate Ocean BGC data under different partitioning schemes.  }
\label{partitioning_fig}
\end{figure}

\subsection{Partitioning Schemes}
Variable relationships and their degree of association often vary across different regions of the spatial domain. Decomposing the spatial domain and applying PCA on individual partition data helps in efficiently modeling the non-linear and complex multivariate relationships using multiple locally linear models~\cite{locallinearembedding, localpca}. 
\clrb{An important distinction to make here is that these partitions are different from the implicit simulation partitions for load-balancing. For a real-world distributed simulation environment, we can partition the individual per-node data to model the overall multivariate data.}
%This also helps in applying our proposed multivariate sampling strategy independently and in parallel across all the partitions. In this paper, we use three different partitioning schemes as discussed below and as shown in Fig.~\ref{partitioning_fig}.

\textbf{(P1) Regular Block-wise Partitioning:} This is a very widely used partitioning scheme where the spatial domain is decomposed into equal sized (non-overlapping) blocks of user-defined dimensions. This scheme does not take into consideration the data properties while segmenting the space. This can lead to sub-optimal partitions for performing local data analysis.

\textbf{(P2) Multivariate K-d Tree Partitioning:} K-d tree is a more data-centric partitioning scheme which follows a top-down sub-division scheme to recursively partition the domain till a desired data properties is met for the partitions~\cite{dutta_pvis17,adr}. 
%Shannon entropy~\cite{Cover2006} is generally the most popular criteria for partitioning the domain using K-d tree scheme. 
In this work, to obtain partitions whose multivariate data can be appropriately modeled using PCA, we design a new terminating \clrr{criterion}. Under this new \clrr{criterion}, we calculate the local PCA of each partition (subsection 3.1). The decision to further sub-divide the partition depends on whether $q$ PC's can capture $p\%$ of the variance in the original multivariate data of the partition. If this criteria is not satisfied for a partition, we further sub-divide the current partition  till either the \clrr{criterion} is met or the size of the partition has reached the minimum dimension set for a partition. The parameters $q$ and $p$ are user-defined and can be appropriately set depending on the application requirements.

\textbf{(P3) Multivariate SLIC Partitioning:} Simple Linear Iterative Clustering (SLIC) is a clustering-based data partitioning scheme~\cite{slic,slic_ucdavis}. 
%It is essentially a variant of k-means clustering algorithm and is commonly used to generate superpixel in images and supervoxels in 3D datasets. 
The partitioning schemes discussed above (\textbf{P1} and \textbf{P2}) produce axis-aligned partitions. SLIC, on the other hand, can produced irregular shaped partition, more in tuned with the data properties. While SLIC based partitioning schemes have been common for univariate data, recently, Jiang et al.~\cite{jiang2018superpca} proposed a method, called superPCA to extend SLIC for multivariate data. They perform SLIC based partitioning on the scalar field of the first PC, which captures maximum variance of the original data. Applying local PCA models on these irregular partitions help us to better model the overall multivariate relationship of the data in a local region.

In Fig.~\ref{partitioning_fig}, we show the results of these partitioning schemes along with the number of PC's required for each partition to capture $99\%$ variance of a 75 variable ocean simulation data (Fig.~\ref{partitioning_fig}a). The number of PC's required for each partition is visualized using a categorical colormap. It can be clearly seen that not all the regions require the same number of PC's to retain the multivariate relationship of the data. Fig.~\ref{partitioning_fig}c and Fig.~\ref{partitioning_fig}d highlight that the data-centric partitioning schemes like \textbf{P2} and \textbf{P3} try to have a more intelligent decomposition of the spatial domain as compared to a regular partitioning scheme. 
%As a result, the local variable relationships can be modeled via PCA in a better way in the corresponding partitions of \textbf{P2} and \textbf{P3} than in \textbf{P1}. 

%Moreover, we can see that the partitions in Fig.~\ref{partitioning_fig}c are of different sizes compare to Fig.~\ref{partitioning_fig}b. Whereas, the partitions of Fig.~\ref{partitioning_fig}d are irregularly shaped. While data-centric schemes like P2 and P3 are good at modeling the data properly, they have the additional overhead of separately storing the partition information and are relatively more time-consuming than simpler schemes like P1. 

To sum up the discussion in this section, for each spatial partition, using our proposed sampling strategy, we reduce the size of the multivariate simulation data and store the reduced form instead of the original high-resolution data. The reduced multivariate data for a partition comprises of the \textit{full PC matrix $\mathbf{C_{full}}$}, \textit{explained variances $\mathbf{EV}$}, the \textit{mean vector $\mathbf{\mu}$}, and the \textit{sampled transformed data $\mathbf{W_s}$} as represented by the orange blocks in Fig.~\ref{overview}.
%The orange boxes in Fig.~\ref{overview} represent this reduced data for each spatial partition.
 
 %To sum up the discussion in this section, for each spatial partition, using a PCA based strategy, we create \textit{multivariate data summaries} that capture the variance of the original data. The multivariate data summary of a partition consists of the PC matrix $\mathbf{C}$, corresponding explained variances $EV$, the mean vector $\mathbf{\mu}$, and  the sampled version of the transformed data $\mathbf{W_s}$.

\section{Multivariate Post-hoc Analysis}
To demonstrate how to perform various multivariate analyses directly on the reduced data, we experimented on two different multivariate data sets. Our first data set is a 2-dimensional ($720 \times 360$), 75-variable ocean biogeochemistry (Ocean-BGC) data set generated using the MPAS-O~\cite{oceanbgc,mpaso,MOORE2001403} (Model for Prediction Across Scale Ocean) and the E3SM~\cite{e3sm} (Energy Exascale Earth System Model) simulations. Our second dataset is a 3-dimensional ($250 \times 250 \times 50$), 13-variable Hurricane Isabel data set, simulating the impact of hurricanes on the coastal regions of the United States. In this section, we cover three different types of multivariate analyses.   

% \usepackage{multirow}

% \begin{table}[]
% \resizebox{\linewidth}{!}{%
% \begin{tabular}{|c|c|c|c|c|c|}
% \hline
% Data Details & \begin{tabular}[c]{@{}c@{}}Partition\\ Scheme\end{tabular} & \begin{tabular}[c]{@{}c@{}}Reduced\\ Size\end{tabular} & \begin{tabular}[c]{@{}c@{}}MV Recon\\ Error\end{tabular} & \begin{tabular}[c]{@{}c@{}}Min Var\\ RMSE\end{tabular} & \begin{tabular}[c]{@{}c@{}}Max Var\\ RMSE\end{tabular} \\ \hline
% \multirow{Ocean BGC\\ Res: 360 x 720\\ Var: 75\\ Size: 416 MB} & Regular & 0.0 & 0.0 & 0.0 & 0.0 \\ \cline{2-6} 
%  & K-d Tree & 0.0 & 0.0 & 0.0 & 0.0 \\ \cline{2-6} 
%  & SLIC & 0.0 & 0.0 & 0.0 & 0.0 \\ \hline
% \multirow{\begin{tabular}[c]{@{}c@{}}Hurricane Isabel\\ Res: 250 x 250 x 50\\ Var: 13\\ Size: 283.4 MB\end{tabular}} & Regular & 0.0 & 0.0 & 0.0 & 0.0 \\ \cline{2-6} 
%  & K-d Tree & 0.0 & 0.0 & 0.0 & 0.0 \\ \cline{2-6} 
%  & SLIC & 0.0 & 0.0 & 0.0 & 0.0 \\ \hline
% \end{tabular}}
% \end{table}

\begin{table}[]
\resizebox{\columnwidth}{!}{%
\begin{tabular}{|c|c|c|c|c|c|}
\hline
\multicolumn{1}{|c|}{Data Details} & \begin{tabular}[c]{@{}c@{}}Partition\\ Scheme\end{tabular} & \begin{tabular}[c]{@{}c@{}}Reduced\\ Size (MB)\end{tabular} & \begin{tabular}[c]{@{}c@{}}Norm MV\\ Recon Error\end{tabular} & \begin{tabular}[c]{@{}c@{}}Min Var\\ RMSE\end{tabular} & \begin{tabular}[c]{@{}c@{}}Max Var\\ RMSE\end{tabular} \\ \hline
Ocean BGC (75 Var) & Regular & 19.8 & 0.00097 & 0.017 & 0.024 \\ \cline{2-6} 
Res: 720 x 360 & K-d Tree & 25.0 & 0.00095 & 0.017 & 0.021 \\ \cline{2-6} 
Size: 416 MB & SLIC & 24.2 & 0.00088 & 0.015 & 0.019 \\ \cline{2-6} \hline
Isabel (13 Var) & Regular & 12.6 & 0.00092 & 0.012 & 0.015 \\ \cline{2-6} 
Res: 250 x 250 x 50 & K-d Tree & 15.1 & 0.00090 & 0.009 & 0.015 \\ \cline{2-6} 
Size: 283.4 MB & SLIC & 15.0 & 0.00085 & 0.007 & 0.010 \\ \cline{2-6} \hline
\end{tabular}}
\\
\caption{Results of data reduction and reconstruction errors for the 3 different partitioning schemes, keeping similar average partition sizes for each datasets and a sampling rate of 5$\%$. }
\label{tab1}
\end{table}
%\vline{0.05}

\textbf{Multivariate Reconstruction:} For the sampled data points, we can reconstruct the full multivariate vector using the reduced data form of the respective partitions. By applying Eq.~\ref{pca_e1} on the transformed data samples $\mathbf{W_s}$, we can reconstruct their original variable values, i.e, $\mathbf{X_s} = \mathbf{W_sC_q} + \mathbf{\mu} $. Since we tried to maximize the variance of the multivariate data (which is inversely proportional to the residual error of reconstruction using Eq.~\ref{pca_e3}), we also get a bound on the multivariate reconstruction error. Table~\ref{tab1} shows the results of data reduction and reconstruction error for the two data sets. For each partition, we store the number of PC's required to capture 99.9$\%$ variance of the data. Using Eq.~\ref{pca_e3}, this is equivalent to a normalized reconstruction error-bound of 0.001 (i.e, $\frac{\epsilon^2}{\| \mathbf{X} \|^2}$ in Section 3.1). This is reflected in the fourth column of Table~\ref{tab1}, where the average normalized reconstruction error across the partitions are under the estimated error-bound of 0.001. 

The normalized root mean square errors (RMSE) of reconstruction vary for the individual variables. Therefore, we report the minimum and the maximum RMSE among the variables in the last two columns of the table. To compare the performance of different partitioning schemes, similar partition sizes were used for all the cases and a sampling rate of 5$\%$ (2.5$\%$ \textbf{S1} and 2.5$\%$ \textbf{S2}) was used to sample the spatial data. It can be seen that for data-centric partitions like K-d tree and SLIC, the average multivariate reconstruction error is lower than regular partitioning. On the other hand, they also tend to have more storage footprint than regular partitions because of the additional overhead of storing partition information.   

\begin{figure}[t!]
\centering
        \includegraphics[width = 0.98\columnwidth]{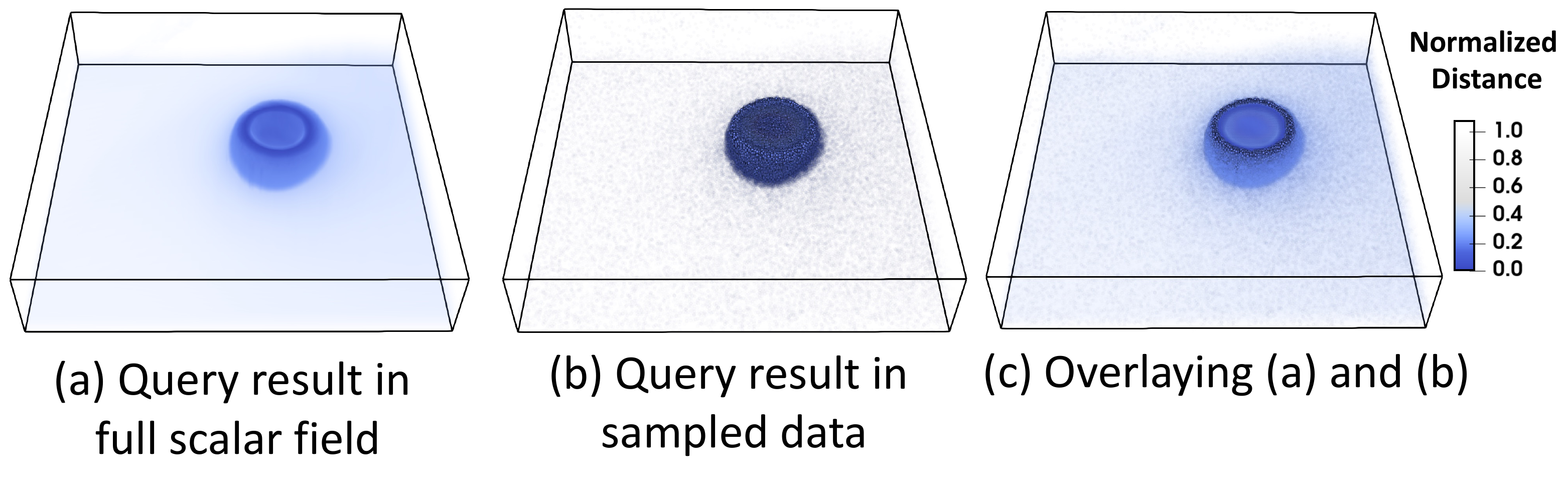} 
\caption{Multivariate query results on Hurricane Isabel data. Query is made for pressure = $-1000 Pa$, temperature = $0 ^{\circ}C$, and wind velocity = $20 ms^{-1}$, which corresponds to the region surrounding the eye of a hurricane. }
\label{mvq_fig}
\end{figure}

%An important point to note here is that, after reconstructing the original multivariate vectors for the samples, the high-resolution scalar fields for the individual variables can also be reconstructed from the sampled data points using various density-estimation and reconstruction algorithms~\cite{}. Given the limited scope of this submission, we do not consider full scalar field reconstruction in this paper. We are focused only on the multivariate reconstruction on the sampled points and performing various analyses using the samples.

\textbf{Multivariate Query-driven Analysis:} Query-driven analysis is a class of very efficient data analysis and visualization methods for large-scale data. 
%Analysis efforts can be targeted only for regions that satisfy an user-provided query of interest. 
For multivariate data, queries are vectors in the original variable space that specify desired values for different variables. Given such a set of multivariate queries $\mathbf{Q}$, we can identify the samples that satisfy $\mathbf{Q}$ without having to reconstruct the multivariate vectors for all the samples. For each partition, we convert $\mathbf{Q}$ to the corresponding orthogonal dimensions using the multivariate data summaries, i.e, $\mathbf{W_Q} = (\mathbf{Q} - \mathbf{\mu})\mathbf{C_q}^T$ (using Eq.~\ref{pca_e2}). We then calculate the distance between these low-dimensional query vectors $\mathbf{W_Q}$ and the transformed samples $\mathbf{W_s}$. By coloring the samples based on the query distance, we can visualize the regions that satisfy the query.

In Fig.~\ref{mvq_fig}, we show the results for a multivariate query of pressure=$-1000 Pa$, temperature=$0 ^{\circ}C$, and wind velocity=$20 ms^{-1}$. This generally corresponds to the wall surrounding the eye of the hurricane. Fig.~\ref{mvq_fig}a shows the multivariate distance of this query on the original multivariate scalar field. Fig.~\ref{mvq_fig}b shows the query distance on the sampled data points. The distance values were normalized for both the cases. In Fig.~\ref{mvq_fig}c, by overlaying the two results, we can qualitatively verify that the query distance measured in the PC space for the samples match with the ground-truth multivariate data.

\textbf{Correlation Analysis:} Understanding the correlation between different variables is a very important analysis objective for multivariate data. Same set of variables can exhibit varying degrees of correlation across different regions of the spatial domain. Many univariate data reduction approaches do not preserve this relationship information in their approaches which can lead to unreliable multivariate analyses. In our proposed method, we store the full PC matrix $\mathbf{C_{full}}$ and their explained variances $\mathbf{EV}$, which are respectively the eigenvectors and the eigenvalues of original covariance matrix. Therefore, the full covariance matrix can be reconstructed using $\mathbf{Cov} = \mathbf{C_{full}}\Lambda_{EV}\mathbf{C_{full}^{-1}}$, where, $\Lambda_{EV}$ is a diagonal matrix whose diagonal elements are the eigenvalues $\mathbf{EV}$. The corresponding correlation matrix (Pearson's \clrr{coefficient}) can then be computed using $\mathbf{Cor = D^{-1} Cov D^{-1}}$, where $\mathbf{D}$ is a diagonal matrix of the standard deviations (square root of the diagonal of $\mathbf{Cov}$).

\begin{figure}[t!]
\centering
        \includegraphics[width = 0.99\columnwidth]{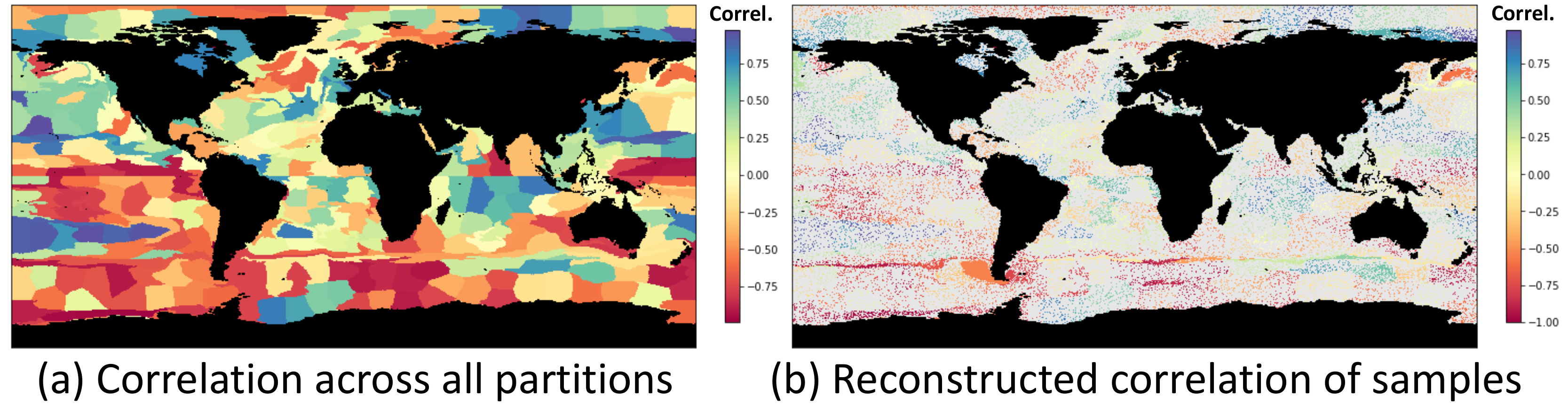} 
\caption{Correlation analysis of \textit{NO$_3$} and \textit{Fe} variables in Ocean BGC dataset}
\label{correl_fig}
\end{figure}

Fig.~\ref{correl_fig} shows the result for correlation analysis between average \textit{NO$_3$} and \textit{Fe} concentration in the Ocean-BGC data set. Fig.~\ref{correl_fig}a shows the original correlation between the two variables for different partitions, created using the SLIC partitioning scheme. The corresponding reconstructed correlation values of the sampled data points are shown in Fig.~\ref{correl_fig}b. By comparing the two figures, we can see that the sampled data-points correctly represent the original correlation information between the two variables.

\section{Evaluation and Discussion}
Apart from the results discussed so far in the paper, we performed different experiments to evaluate our proposed multivariate sampling strategy and to understand the impact of different factor. Here is a brief outline of our observations from these experiments: 

\textbf{Effect of Partition Size:} Partition size plays a crucial role in modeling the multivariate relationship. We observed that for a particular partitioning scheme, the average variable reconstruction error increases with increasing partition sizes. This is mainly because, for bigger spatial partitions, the variable relationships are often complex and non-linear. As a result, linear models like PCA are not good at capturing their multivariate properties. 
%For relatively smaller partition sizes, by using the linear PCA model for each partition, we are able to model the overall non-linear multivariate relationship across the spatial domain in a much better manner.    

\textbf{Effect of Partition Schemes:} Not only the size, but also the shape of the partition plays an important role in modeling the multivariate data using local PCAs. Out of the three partitioning schemes that we used in this work, we found that for similar partition sizes, \textbf{P3} performs the best, followed by \textbf{P2} and \textbf{P1}. 
%This is because both \textbf{P3} and \textbf{P2} decide the partition shape based on multivariate data properties, unlike \textbf{P1}. 
\clrb{However, there is a trade-off with respect to the size of the data summaries as well as the computation time for data-centric partitions like \textbf{P3} and \textbf{P2} against simpler schemes like \textbf{P1}. Moreover, for data-centric schemes the shape of the partitions will have to be updated for every timestep in the simulation. Based on these observations, we feel that if we apply our strategy in an in-situ scenarios, \textbf{P1} would be an ideal candidate because of its simplicity and less computational overhead. On the other hand, for small scale offline analyses without much computational constraints, \textbf{P3} and \textbf{P2} can better model the overall non-linear multivariate relationships.}

\textbf{Effect of Sample Rate and Algorithm:} The rate of sampling the data using different sampling algorithms also play a important role in the quality of post-hoc analysis as well as the overall storage footprint of the data summaries. More samples essentially help get better reconstruction results, however at the cost of more storage size. The sampling algorithm can also effect the quality of the analysis results. Among the two sampling algorithms used in the paper, \textbf{S1} preserves the overall data distribution, whereas, \textbf{S2} is more tailored towards preserving the important features in the data. Depending on the requirement of analysis, users can decide a flavor of sampling algorithm for their implementation, or even combine the results of multiple algorithms. 

%It is also possible to combine the features of different algorithms by taking fractions of samples from different algorithms. For this work, we used 50$\%$ of the total sampling rate from \textbf{S1} and remaining 50$\%$ from \textbf{S2}.

\section{Conclusion and Future work}

In this paper, we have proposed a sampling strategy which utilizes and preserves the variables relationships, while reducing the storage footprint of multivariate simulation data. The global multivariate relationship is captured using multiple local PCA models across the spatial domain, which is decomposed using different multivariate relationship-aware partitioning schemes. We showed how various multivariate analysis and visualization tasks can be performed directly on the sampled data without the need to reconstruct the high-resolution scalar fields. \clrb{In future, we plan to extend this strategy for temporal multivariate data sets by using incremental PCA models. We also plan to deploy this sampling strategy for in-situ data reduction of multivariate data in large--scale simulation models.}

%In future, we would like to extend this strategy to temporal multivariate data sets by using incremental/streaming PCA models. As simulations produce data at every timestep, we can incrementally update the local PCA models to further reduce the storage footprint as well as model the variable relationships across time. After this extension, we plan to deploy this sampling strategy to a real-world in situ simulation setup. We also plan to further investigate into different partitioning strategies to efficiently model multivariate data. We hope to incorporate these ideas into the proposed approach and along with more evaluations submit as a full journal paper in the near future.   

\acknowledgments{
This research was supported by the Laboratory Directed Research and Development program of Los Alamos National Laboratory under project number 20200065DR.  PJW was supported as part of the Energy Exascale Earth System Model (E3SM) project, funded by the U.S. Department of Energy, Office of Science, Office of Biological and Environmental Research.(LA-UR-20-24913).
}

\bibliographystyle{abbrv-doi}

\bibliography{template}
\end{document}